\documentclass[twoside,11pt]{article}

\usepackage{jmlr2e}
\usepackage{amsmath}
\usepackage{algorithm2e}
\usepackage{booktabs}
\usepackage{color}

\SetKwInput{KwInput}{Input}
\SetKwInput{KwOutput}{Output}

\ShortHeadings{Application of SVM+ in Early Drug Discovery}{Niharika, Lars and Ola}
\firstpageno{1}

\begin{document}

\title{Conformal Prediction in Learning Under Privileged Information Paradigm with Applications in Drug Discovery}

\author{\name Niharika Gauraha \email niharika.gauraha@farmbio.uu.se \\
       \addr Department of Pharmaceutical Biosciences \\
       Uppsala University\\
       Uppsala, Sweden 
       \AND
       \name Lars Carlsson \email  Lars.A.Carlsson@astrazeneca.com \\
       \addr Quantitative Biology, Discovery Sciences,\\ IMED Biotech Unit, AstraZeneca\\
       Gothenberg, Sweden
       \AND
       \name Ola Spjuth  \email Ola.Spjuth@farmbio.uu.se \\
       \addr Department of Pharmaceutical Biosciences \\
       Uppsala University\\
       Uppsala, Sweden
       }

\editor{}

\maketitle

\begin{abstract}
This paper explores conformal prediction in the learning under privileged information (LUPI) paradigm. We use the SVM+ realization of LUPI in an inductive conformal predictor, and apply it to the MNIST benchmark dataset and three datasets in drug discovery. The results show that using privileged information produces valid models and improves efficiency compared to standard SVM, however the improvement varies between the tested datasets and is not substantial in the drug discovery applications. More importantly, using SVM+ in a conformal prediction framework enables valid prediction intervals at specified significance levels.



\end{abstract}


\section{Introduction}
The growing availability of data offers great opportunities but also many challenges to develop models which can be used to make predictions about future observations. The classical machine learning paradigm is: given a set of training examples in the form of iid pairs

\begin{equation}\label{eq:def_nonconformity}
(x_1, y_1), ..., (x_l, y_l), \hspace{1em} x_i \in X, \hspace{1em} y_i \in \{-1, +1\}
\end{equation}

\noindent seek a function that approximates the unknown decision rule in the best possible way and provides the smallest probability of incorrect classifications. Training examples are represented as features $x_i$ and the same feature space is required for predicting future observations. However this approach does not make use of other useful data that is only available at training time; such data is referred to as Privileged Information (PI) \citep{Vapnik:2009kl}. Hence much data that could improve models is set aside and not included in the training process.

In the Learning Using Privileged Information (LUPI) paradigm, training examples instead come in the form of iid triplets

\begin{equation}\label{eq:def_nonconformity}
(x_1,x_1^*, y_1), ..., (x_l,x_l^*, y_l), \hspace{1em} x_i \in X, \hspace{1em} x^*_i \in X^*, \hspace{1em} y_i \in \{-1, +1\}
\end{equation}

\noindent where $x^*$ denotes PI. The objective is the same as in classical machine learning, with the extension that privileged information is available in the training stage. One implementation of LUPI is SVM+, and \cite{Vapnik:2009kl} showed that this approach and implementation can accelerate the learning process, and outperform classical machine learning in a set of applications.

Conformal prediction~\citep{vovk2005algorithmic} is a method that provides a layer on top of an existing machine learning method and uses available data to determine valid prediction regions for new examples. In contrast to standard machine learning that delivers point estimates, conformal prediction yields a prediction region that contains the true value with probability equal to or higher than a predefined level of confidence. Such a prediction region can be obtained under the assumption that the observed data is exchangeable.

In this work we explore conformal prediction in the LUPI paradigm with the aim to improve predictive performance and obtain valid prediction regions. We study the effects of the SVM+ realization of LUPI in an inductive conformal predictor on a benchmark dataset and provide examples in drug discovery problems where machine learning has become a core part of the early discovery process~\citep{Norinder:2014bf,bendtsen2017improving,zhang2017machine}.



\section{Data and Methods}
\subsection{Support Vector Machines (SVM)}
Support vector machines~\citep{vapnik1998statistical}, are one of the most successful methods for classification in machine learning. 
One of the key concepts of SVM is the use of separating hyperplanes to define decision boundaries, and the optimal decision hyperplane is a plane in a multidimensional space that separates between data points of different classes and that also maximizes the margin, separating the two classes. SVM uses the kernel trick to generate a high dimensional nonlinear representation of the input data examples where it performs the separation with a continuous separation hyperplane, such that the distances of misclassified data examples from the hyperplane are minimized. In this study, we use a classification SVM for training our classification models with a Radial Basis Function (RBF) kernel
 \begin{align*}
{\displaystyle K(x_i ,x_j )=\exp(-\gamma \|x_i -x_j \|^{2})}, 
\end{align*}
where $\gamma$ controls the width of the kernel function, and $x_i$ and $x_j$ are the vectors of the $i$th and the $j$th training samples, respectively. The kernel parameters $\gamma$ and the SVM cost parameter $C$ are tuned using two-dimensional cross-validated grid search.

\subsection{SVM+}
Realizations of LUPI~\citep{Vapnik:2009kl} are mostly based on SVM and referred to as SVM+. In SVM+, the privileged information (PI) is used to estimate the slack variables, which are defined as the distance between the support vectors and the decision boundary. The PI provides a means for regularizing the SVM optimization problem and assists in its generalization. 
This can be also viewed as augmenting the standard SVM with a second kernel that defines a similarity measure between any two data points in a privileged information space. We use RBF for both kernels, where the first kernel parameters are tuned using SVM on $X$ (standard features) and the second kernel parameter is tuned using using SVM on $X^*$ (PI).

\subsection{Conformal prediction}
Traditional machine learning algorithms for classification problems simply predicts the class labels without any confidence. Conformal predictors expand on this as they output prediction regions for a specific confidence level provided by the user. The confidence value is an indication of how likely each prediction is of being correct, for example, a confidence of 95\% implies that the percentage prediction error will be 5\% on average. Conformal predictors are built on top of traditional machine learning algorithms, referred to as underlying algorithms, and they can be broadly categorized into transductive and inductive approaches; we refer to \cite{papadopoulos2008inductive} for more details. We here consider the inductive approach called Inductive Conformal Prediction (ICP), which is more computationally efficient as compared with the transductive approach. In particular, we use Mondrian ICP 
with SVM or SVM+ as the underlying algorithms, and the SVM or SVM+ distance to the decision boundary to define the non-conformity measures (NCM).
Mondrian conformal prediction has the advantage that we achieve validity for the individual classes.
To evaluate the performance of conformal predictors, we consider the observed fuzziness, as defined in \cite{vovk2005algorithmic}.

\subsection{Data}
As a reference dataset we used the MNIST dataset~\citep{LeCun:1998gd}, which has been used previously with the SVM+ algorithm~\citep{Vapnik:2009kl}. The MNIST dataset contains grayscale images of handwritten digits  0-9 as vectors of 28 x 28 pixel images, and was downloaded from \url{http://yann.lecun.com/exdb/mnist/}. 
We used a 4000 example subset of MNIST dataset comprising digits 5 and 8. The original $28 \times 28$ pixel images were used as PI, where images resized to $8 \times 8$ pixel resolution were used as standard dataset.
We also used three datasets in drug discovery (Hansen, MMP, and AHR), where chemical structures are represented as numerical features, and the response variable is measured in a biological assay. 
The Hansen dataset~\citep{Hansen:2009kq} was constructed to enable the prediction of mutagenicity for the chemical structure of \textit{e.g.} a drug candidate, based on measurements from the Ames Mutagenicity test~\citep{Zeiger:2001wd}. 
The MMP dataset is based on measurements for small molecule disruptors of the Mitochondrial Membrane Potential, and is commonly used to assess the effect of chemicals on mitochondrial function~\citep{Sakamuru:2016fv}.
The AHR dataset is based on measurements for interaction with the aryl hydrocarbon receptor (AHR), related to chemical toxicity and interaction with drugs and other compounds~\citep{Bradshaw:2009jk}.
AHR and MMP were downloaded from PubChem (AHR PubChem AID: 743122, MMP PubChem AID: 720637) as part of the Tox21 project that has previously been used for modeling~\citep{Huang:2016lq}. The Hansen dataset was downloaded from \url{http://doc.ml.tu-berlin.de/toxbenchmark/}.
The chemical structures in the datasets AHR, MMP and Hansen were represented using ten Physical-Chemical descriptors (Chi1n,  Chi2n,  Chi3n,  Chi4n,  Chi0v,  C
hi1v,   Chi2v,  Chi3v,  Chi4v and MolLogP),
and Morgan fingerprints calculated using RDKit (www.rdkit.org). The Physical-Chemical descriptors contains less features and can be hypothesized to produce less accurate models than Morgan fingerprints.
All the datasets are binary class problems with class labels (-1, 1). The details of the datasets are given in Table~\ref{tab:datasets}.

\begin{table} [h!] 
\caption{Description of the datasets and feature sets used in this work} \centering
\vspace{1em}
\begin{tabular}{llllc}
\toprule
Dataset & Features & & \# Observations & \# Features \\
\midrule
MNIST & X: \hspace{0.5em} $8 \times 8$ pixel images && 4000 & 64 \\
& X*: $28 \times 28$ pixel images && 4000 & 784   \\
\hline
AHR & X: \hspace{0.5em}Phys-chem descriptors && 6299 & 10 \\
& X*: Morgan fingerprints && 6299 &  55725 \\
\hline
Hansen & X: \hspace{0.5em}Phys-chem descriptors && 6509 & 10 \\
& X*: Morgan fingerprints && 6509 & 48325  \\
\hline
MMP & X: \hspace{0.5em}Phys-chem descriptors && 5647 & 10 \\
& X*: Morgan fingerprints && 5647 & 49764 \\
\bottomrule
\end{tabular}\label{tab:datasets}
\end{table}

\subsection{Study design}
We denote $X = (x_1,...,x_l)^T$, as a matrix of standard features, $X^* = (x^*_1,...,x^*_l)^T$, as a matrix of PI, and $y = (y_1,...,y_l)$, as a vector of class labels. We chose three statistical models: SVM on $X$, SVM on $X^*$ and SVM+ on $X$ with $X^*$ as PI. These three models were applied on all four datasets to compare their predictive accuracy and efficiency (observed fuzziness). First, the dataset was partitioned using stratified-split into two parts: training (80\%) and external test (20\%) set, and the training and the test sets were then kept fixed. Then the training set was randomly divided into proper-training (70\%) and a calibration set (30\%).
For tuning the parameters for each model and for each dataset, we used five-fold cross validation technique on the corresponding proper-training set, and we selected the parameters based on the highest prediction accuracy. More importantly, the tuning was performed in three steps:\\

\begin{enumerate}
\item Tuning of the first RBF kernel parameter, $\gamma_1$, and the SVM parameter, $C_1$, for the model SVM on $X$: We used two dimensional grid search with 5-fold CV on the $X$-proper-training set. 
\item Tuning of the second RBF kernel parameter, $\gamma_2$, and the SVM parameter, $C_2$, for the model SVM on $X^*$: We used two dimensional grid search with 5-fold CV on the $X^*$-proper-training set.
\item Tuning of SVM+ parameters, $C$ and $\gamma$: We used two dimensional grid search with 5-fold CV on the $X$-proper-training and $X^*$-proper-training with selected kernel parameters, $\gamma_1$ and $\gamma_2$, in the previous steps.
\end{enumerate}

\begin{table} [h!]
\caption{Hyper parameter ranges for various methods} \label{tab:param}
\centering
\begin{tabular}{llllc}
\hline
Method & & C & & $\gamma$ \\
\hline
SVM on $X$ & &  [.1, 1000] &&  [1e-7, 1] \\
SVM on $X*$ & &  [.1, 1000] &&  [1e-7, 1] \\
SVM+ on $X$ (with $X^*$ as PI) & &  [.01, 100] &&  [1e-4, .1] \\
\hline
\end{tabular}
\end{table}

The ranges explored for each parameter and for each method are given in Table~\ref{tab:param}. 
The proper-training set with corresponding selected parameters was then used to build the model. The non-conformity scores were computed on the corresponding calibration set. We used the SVM/SVM+ decision function to define the non-conformity measure (NCM)
\begin{align*}
	\alpha_i = y_if(x_i ),
\end{align*}
where $f$ is the SVM/SVM+ decision function.
Then for each observation in the external test set, we computed (Mondrian) conformal prediction p-values for each class. The above procedure was repeated 10 times, and the average predictive performance, and the average observed fuzziness was reported.
The above-mentioned steps are outlined in Algorithm~\ref{alg:studydesign}.

\begin{center}
\begin{algorithm}[h!] 
\fbox{\parbox{\hsize}{

 \KwInput{$(X, \;X^*,\; y)$, N: number of repetitions}
 \KwOutput{average prediction accuracy, average validity, average observed fuzziness}
  Step 1: Partition the dataset $(X, \;X^*,\; y)$ into 80\% for training, and 20\% for testing using stratified split.\\
  Step 2: Partition the training set, $(X, \;X^*,\; y)$-training, into 70\% proper-training and 30\% calibration. \\
  Step 3: Use cross-validation for tuning $C_1$ and $\gamma_1$, using SVM on $X$-proper-training, and select the one that gives the highest average prediction accuracy.\\
  Step 4: Use cross-validation for tuning $C_2$ and $\gamma_2$, using SVM on $X^*$-proper-training, and select the one that gives the highest average prediction accuracy.\\
  Step 5: Use cross-validation for tuning $C$ and $\gamma$, using SVM+ on $X$-proper-training with $X^*$-proper-training as PI, and the kernel parameters $\gamma_1$ and $\gamma_2$ as selected in Step 3 and Step 4 respectively.\\
 Step 6:\\
\Repeat{ N iterations}{
  Step 6.1: Randomly partition the $(X, \;X^*,\; y)$-training set into 70\% for proper-training, and 30\% for calibration. \\
  Step 6.2: Train the three models using their corresponding proper-training set.\\
  Step 6.3: Compute NCM using the corresponding calibration set for each model.\\
  Then for each model compute the following on their corresponding test set:\\
    - prediction accuracy \\
  - deviation from exact validity \\
  - observed fuzziness \\
 }
 Return average prediction accuracy, average validity and average observed fuzziness of each model.
 }}
 \caption{Algorithm for study design: ICP with SVM and SVM+}
 \label{alg:studydesign}
 \end{algorithm}
\end{center}

\subsection{Computational Details}
The computations were performed on resources provided by SNIC through Uppsala Multidisciplinary Center for Advanced Computational Science (UPPMAX) under Project SNIC 2017-7-273. We used existing Python implementation of LibSVM in scikit-learn toolkit for training and prediction SVM models. We implemented SVM+ on Python using python-cvxopt: Python package for convex optimization. Conformal prediction using SVM as an underlying machine learning algorithm was implemented in Python using scikit-learn toolkit.

\section{Results and Discussion}
\begin{figure}[h!]\centering
	\includegraphics[scale=.5]{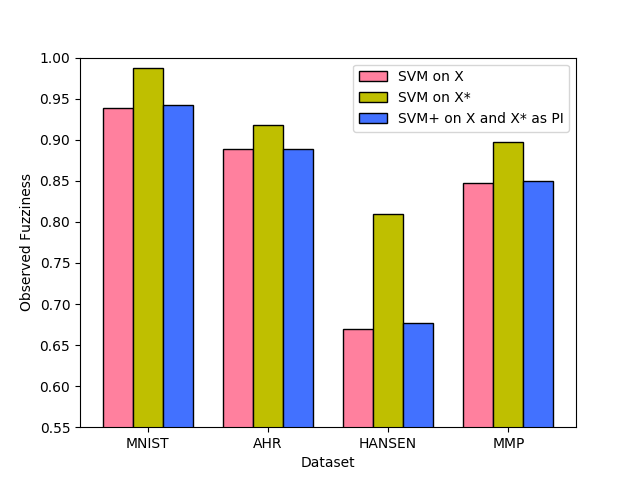}
\caption{Comparision of prediction accuracy on four selected datasets using SVM on X (pink), SVM on X*(yellow) and SVM+ (blue).} 
\label{fig:pred-performance}
\end{figure}

In this study, we have used prediction accuracy and observed fuzziness as measures of performance.
The prediction accuracies of the three statistical models are given in Table \ref{tab:pred-performance} and in Figure \ref{fig:pred-performance}, and we observe that the methodology using SVM on X* outperforms the other models in terms of prediction accuracy for all datasets, but we note that SVM+ outperforms SVM on X for most of the datasets.

\begin{figure}\centering
	\includegraphics[scale=.5]{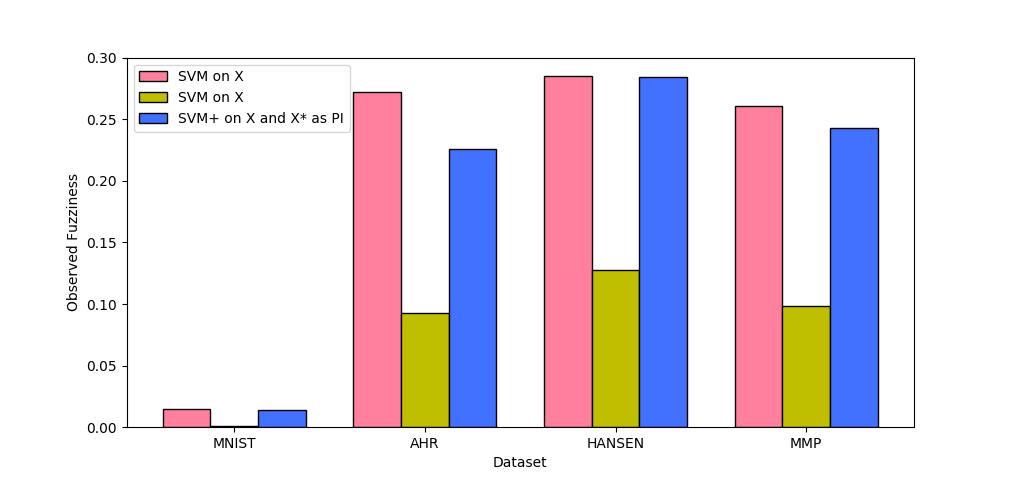}
	\caption{Comparision of observed fuzziness on four selected datasets using SVM on X (pink), SVM on X*(yellow) and SVM+ (blue).
	}\label{fig:valEff}
\end{figure}

For comparison of the three Mondrian inductive conformal predictors, their measure of efficiency and validity are given in Table \ref{tab:valEff} and in Figure~\ref{fig:valEff}. The smaller the efficiency (observed fuzziness) is, the better the model performs. Also here we see that SVM on X* performs best in terms of efficiency, and that SVM on X* outperforms SVM on X for all datasets, but that the level of improved efficiency with SVM+ varies between the datasets.

One implication of using SVM+ is the need for tuning additional SVM hyper-parameters associated with PI, which increases the computational complexity substantially.

\begin{table} [h!]
\caption{Comparison of prediction accuracy}\label{tab:pred-performance}
\centering
\begin{tabular}{lllcc}
\toprule
Dataset &  Statistical Model && prediction accuracy\\
\midrule
MNIST & SVM on $X$ && 0.939125 \\
&SVM on $X^*$ &&  0.987375 \\
&SVM+ on $X$ with $X^*$ as  PI && 0.942875 \\
\hline
AHR & SVM on $X$ && 0.888889 \\
&SVM on $X^*$ &&  0.917857 \\
&SVM+ on $X$ with $X^*$ as  PI && 0.888889 \\
\hline
Hansen & SVM on $X$ && 0.669124 \\
&SVM on $X^*$ &&  0.809370 \\
&SVM+ on $X$ with $X^*$ as  PI && 0.676651 \\
\hline
MMP & SVM on $X$ && 0.847522 \\
&SVM on $X^*$ &&  0.896726 \\
&SVM+ on $X$ with $X^*$ as  PI && 0.849292 \\
\bottomrule
\end{tabular}
\end{table}

\begin{table} [h!]
\caption{Comparision of validity and efficiency}\label{tab:valEff}
\centering
\begin{tabular}{lllcc}
\toprule
Dataset & & Learning Algorithm & Validity & observed fuzziness \\
\midrule
MNIST  && SVM on $X$& 0.189254 & 0.015103\\
 && SVM on $X^*$ & 0.182684& 0.000839\\
&& SVM+ on $X$ with $X^*$ as  PI &   0.176197 & 0.013733 \\
\hline
AHR  && SVM on $X$& 0.168761 & 0.272146\\
 && SVM on $X^*$ & 0.107204& 0.092754\\
&& SVM+ on $X$ with $X^*$ as  PI &   0.100159 & 0.226047 \\
\hline
Hansen  && SVM on $X$& 0.121286 & 0.285467\\
 && SVM on $X^*$ &  0.128802 & 0.127245  \\
&& SVM+ on $X$ with $X^*$ as  PI & 0.130737 &  0.283943 \\
\hline
MMP && SVM on $X$& 0.164847 &  0.260612\\
 && SVM on $X^*$ & 0.140734 & 0.098628 \\
&& SVM+ on $X$ with $X^*$ as  PI & 0.151952 & 0.248843  \\
\bottomrule
\end{tabular}
\end{table}

\section{Conclusions}
We here introduced conformal prediction using LUPI/SVM+ as underlying method. We investigated the validity and efficiency of inductive conformal predictors with SVM+ on the MNIST benchmark dataset, and also applied it to three datasets relevant to drug discovery. Our results show that the ICP with SVM+ is more efficient than ICP with SVM on X, in terms of observed fuzziness. We also showed that the prediction accuracy of SVM+ on X with X* as privileged information is better than standard SVM on X for all datasets, however in some cases the improvements on observed fuzziness and prediction accuracy are only marginal.

\newpage
\acks{We would like to acknowledge Alexander Kensert and Jonathan Alvarsson for assistance in data preparation. This project received financial support from the Swedish Foundation for Strategic Research (SSF) as part of the HASTE project under the call 'Big Data and Computational Science'.
The computations were performed on resources provided by SNIC through Uppsala Multidisciplinary Center for Advanced Computational Science (UPPMAX) under project SNIC 2017/7-273.}

%
%
%

\vskip 0.2in
\bibliography{svmplus-cp}

\end{document}